\documentclass{article}

\usepackage{PRIMEarxiv}

\usepackage[utf8]{inputenc} 
\usepackage[T1]{fontenc}    
\usepackage{hyperref}       
\usepackage{url}            
\usepackage{booktabs}       
\usepackage{amsfonts}       
\usepackage{nicefrac}       
\usepackage{microtype}      
\usepackage{lipsum}
\usepackage{fancyhdr}       
\usepackage{graphicx}       
\usepackage{enumitem}
\usepackage{subcaption}
\usepackage{makecell}
\usepackage{amsmath}
\usepackage{pifont}

\graphicspath{{media/}}     

\title{ArthroPhase: A Novel Dataset and Method for Phase Recognition in Arthroscopic Video
\thanks{
\textbf{This is a preprint of a manuscript submitted to Computer Assisted Surgery. The final version may be different after peer review and acceptance.}} 
}

\author{
  Ali Bahari Malayeri, Matthias Seibold, Nicola Cavalcanti, Jonas Hein, Sascha Jecklin \\
  Research in Orthopedic Computer Science (ROCS)\\
  Balgrist University Hospital, University of Zurich \\
  Zürich, Switzerland \\
  \texttt{Corresponding author: ali.baharimalayeri@balgrist.ch} \\
   \And
  Lazaros Vlachopoulos, Sandro Fucentese, Sandro Hodel \\
  Department of Orthopedics \\
  Balgrist University Hospital, University of Zurich \\
  Zürich, Switzerland \\
   \And
  Philipp Fürnstahl \\
  Research in Orthopedic Computer Science (ROCS)\\
  Balgrist University Hospital, University of Zurich \\
  Zürich, Switzerland \\
}

\begin{document}
\maketitle

\begin{abstract}
\textbf{Purpose} \\
This study aims to advance surgical phase recognition in arthroscopic procedures, specifically Anterior Cruciate Ligament (ACL) reconstruction, by introducing the first arthroscopy dataset and developing a novel transformer-based model. We aim to establish a benchmark for arthroscopic surgical phase recognition by leveraging spatio-temporal features to address the specific challenges of arthroscopic videos including limited field of view, occlusions, and visual distortions.

\textbf{Methods} \\
We developed the ACL27 dataset, comprising 27 videos of ACL surgeries, each labeled with surgical phases. Our model employs a transformer-based architecture, utilizing temporal-aware frame-wise feature extraction through a ResNet-50 and transformer layers. This approach integrates spatio-temporal features and introduces a Surgical Progress Index (SPI) to quantify surgery progression. The model's performance was evaluated using accuracy, precision, recall, and Jaccard Index on the ACL27 and Cholec80 datasets.

\textbf{Results} \\
The proposed model achieved an overall accuracy of 72.91\% on the ACL27 dataset, with a precision of 72.86\% and Jaccard Index of 57.39\%. On the Cholec80 dataset, the model achieved a comparable performance with the state-of-the-art methods with an accuracy of 92.4\%, precision of 85.6\%, recall of 80.6\%. The SPI demonstrated an output error of 10.6\% and 9.86\% on ACL27 and Cholec80 datasets respectively, indicating reliable surgery progression estimation.

\textbf{Conclusion} \\
This study introduces a significant advancement in surgical phase recognition for arthroscopy, providing a comprehensive dataset and a robust transformer-based model. The results validate the model's effectiveness and generalizability, highlighting its potential to improve surgical training, real-time assistance, and operational efficiency in orthopedic surgery. The publicly available dataset and code will facilitate future research and development in this critical field.\end{abstract}

\keywords{Surgical phase recognition \and orthopedics \and arthroscopy \and transformer}

\section{Introduction}
In the field of surgical interventions, automated phase recognition and analysis enable a wide range of downstream applications, i.e. detecting complications, optimise clinical workflows, documenting procedures, and generating automated surgery reports, thus emphasizing the critical importance of surgical phase recognition \cite{1.maier2017surgical}. Furthermore, understanding the surgical context is a fundamental task for the development of advanced computer-aided surgery systems that enable intelligent surgical assistance. By leveraging various data modalities such as signals \cite{21.padoy2012statistical} and video \cite{2.twinanda2016endonet}, particularly endoscopic video as one of the most promising data sources, machine learning and advanced algorithmic approaches have significantly enhanced the capabilities of surgical systems \cite{deo2015machine}.

Automated surgical workflow recognition for clinical applications such as cholecystectomy, cataract removal, and rectal resection have been extensively investigated in previous work \cite{demir2023deep}. While there has been some attempts in summarizing arthroscopic videos, such as the work by Lux et al. on keyframe-based video summarization \cite{lux2010novel}, comprehensive surgical workflow recognition for arthroscopic procedures remains largely unexplored. This gap underscores the importance and drives our research in automatic phase detection for arthroscopy. Surgical phase recognition in arthroscopy presents several unique challenges, including blurry frames, scenes obscured by liquid, floating particles, limited workspace, and inherent ambiguity in individual frames. Our approach addresses these challenges by leveraging a transformer-based model and introducing a novel dataset specifically for arthroscopic surgical phase recognition, aiming to establish a scientific benchmark and improve the accuracy and applicability of phase recognition in this domain.

The problem of surgical phase recognition has initially been tackled using methods like Hidden Markov Model (HMM) that operated on manually crafted binary signals from the OR, such as synchronized signals acquired over time, thereby setting foundational techniques in the field \cite{21.padoy2012statistical}. Later, frame-based analysis using deep learning models, particularly image-based Convolutional Neural Networks (CNNs), marked a significant advancement by classifying individual frames based on spatial features \cite{2.twinanda2016endonet}; subsequently, spatio-temporal methods were developed to not only improve feature extraction but also explicitly model the dependencies between frames in a sequence, a critical aspect in surgical video analysis, with approaches such as Long-Short-Term Memory (LSTM) \cite{17_10.1162/neco.1997.9.8.1735} and Temporal Convolutional Networks (TCN) \cite{13.lea2017temporal}, significantly enhancing the ability to capture long-range dependencies within surgical videos. More recently, Transformer-based methods have revolutionized the approach by efficiently managing these dependencies across extensive sequences using self-attention mechanisms and positional embedding \cite{14.vaswani2017attention}, allowing to handle much longer sequences.

This study presents the first comprehensive investigation into surgical phase recognition in the field of arthroscopy, introducing a novel dataset consisting of 27 videos captured from ACL surgeries including surgical phase labels. We propose a novel method that sets a scientific benchmark in the field of surgical phase recognition for arthroscopic procedures. Our approach utilizes a transformer-based model to tackle the challenges of surgical phase recognition in arthroscopy. We employ temporal-aware frame-based features essential for accurately analyzing and predicting phases in long surgical videos. These spatio-temporally coherent features are fed into two branches of transformers that differ in input length, enhancing temporal resolution and precision in identifying and classifying surgical phases. In addition to phase recognition, our model incorporates a novel parallel output termed the Surgical Progress Index (SPI), which provides a straightforward measure for the progression of the surgical procedures. The associated code and dataset will be made publicly available upon acceptance. The contributions of this work can be summarized as follows:

\begin{itemize}[leftmargin=0.8cm]
    \item We establish a foundational scientific benchmark for surgical phase recognition in arthroscopy by providing a novel publicly available dataset consisting of 27 Anterior Cruciate Ligament surgery videos with surgical phase labels.
    \item We propose an effective mechanism, the Surgical Progress Index (SPI), which implicitly integrates global temporal context and provides a valuable measure for the progress of a surgery.
    \item We propose the utilization of spatio-temporal features, provide a comprehensive ablation study and benchmark the proposed method with tate-of-the-art approaches from other domains like laparoscopy.
\end{itemize}

\section{State-of-the-art}
In the early stages of surgical phase recognition, manually recorded data on tool usage and equipment were used to interpret and classify different segments of surgery. Algorithms like Dynamic Time Warping and Hidden Markov Models were applied to process this data, particularly in cholecystectomy surgeries \cite{21.padoy2012statistical}. These methods aimed to effectively segment and recognize workflow phases \cite{21.padoy2012statistical}. Automated surgical phase recognition evolved significantly with the introduction of EndoNet \cite{2.twinanda2016endonet}, the first model to utilize Convolutional Neural Networks (CNN) for extracting spatial features from video data. This approach was further developed by integrating these features into a hierarchical Hidden Markov Model to capture temporal dynamics, establishing the groundwork for multi-task recognition within cholecystectomy surgery videos. Enhancing the foundational approach of EndoNet, Twinanda \textit{et al.} \cite{Twinanda2016SingleAM} substituted the hierarchical HMM with LSTM networks to deepen the model's understanding of temporal sequences, marking a pivotal shift towards intricate temporal analysis. This advancement paved the way for the SV-RCNet framework, which harnessed the combined power of ResNet and LSTM to holistically capture spatial and temporal features, marking a significant improvement in spatio-temporal feature representation within an end-to-end learning framework \cite{10_jin2017sv}. Up to this point, most papers used Cholec80, Cholec120, and M2CAI16 datasets which contain laparoscopic videos of cholecystectomy surgeries. Subsequently, a video dataset named Cataract-101 \cite{x1_schoeffmann2018cataract}, containing recordings of 101 cataract surgeries, was published. Yu \textit{et al.} \cite{x2_yu2019assessment} leveraged the Cataract-101 dataset, employing various deep transfer learning algorithms to classify surgery phases in cataract surgery videos, inspiring further research for this clinical application. The advancement of surgical phase recognition continued with the development of MTRCNet-CL by Jin \textit{et al.} \cite{18_jin2020multi}, which incorporated a correlation loss to leveraging the relationship between tool detection and phase recognition using Cholec80, thereby enhancing the performance of both tasks simultaneously. This model introduces the strategic use of multiple loss terms to refine learning outcomes. Addressing the limitations of LSTM in handling long sequences, Czempiel \textit{et al.} \cite{22_czempiel2020tecno} proposed TeCNO, which utilizes Temporal Convolutional Networks (TCN) for enhanced temporal feature integration in surgical videos. This method marks a significant improvement in the continuous and dynamic recognition of surgical phases, refining phase detection capabilities in complex video sequences.

Building on these achievements, the introduction of Transformer models \cite{21_vaswani2017attention}, originally developed for Natural Language Processing (NLP), has significantly advanced surgical video analysis. Their inherent ability to analyze elements of sequences in parallel, without the constraints of sequential processing, makes them particularly suited for maintaining temporal context in the extended sequences typical of surgical video, thus aiding in precise phase recognition. OperA \cite{czempiel2021opera} by Czempiel \textit{et al.} utilizes a transformer-based approach with attention regularization to significantly enhance surgical phase recognition. Trans-SVNet introduced by Gao \textit{et al.} \cite{gao2021trans} represents another significant advancement in Transformer application for surgical phase recognition. This model uniquely combines ResNet and TCN to generate spatial and temporal embeddings, which are then effectively aggregated using a Transformer framework to enhance phase recognition accuracy. Liu \textit{et al.} \cite{liu2023lovit} introduced LoViT, which enhances online surgical phase recognition by integrating a temporally-rich spatial feature extractor with a multi-scale temporal aggregator. Utilizing self-attention and ProbSparse mechanisms, LoViT efficiently processes local and global features, achieving superior performance and robustness compared to state-of-the-art methods. Additionally, Liu \textit{et al.} \cite{liu2023skit} developed SKiT, a model that uses an efficient key pooling operation to capture crucial information in lengthy surgical videos, achieving higher accuracy and constant inference time, making it suitable for real-time recognition systems. Holm \textit{et al.} \cite{holm2023dynamic} further expanded the scope of surgical phase recognition by employing dynamic scene graphs to represent surgical videos, enhancing the robustness and explainability of model predictions.

\section{Methods}
\subsection{Development and Annotation of a Comprehensive ACL Surgery Dataset}
We introduce a unique dataset of arthroscopic video collected in our institution, using a Karl Storz Laparoscope. Specifically, we identified 27 surgery recordings focusing on Anterior Cruciate Ligament (ACL) reconstruction, collected from January 2022 until June 2023, that met our inclusion criteria: they exclusively include standard ACL procedural steps and involve cases without any additional pathologies requiring further surgical interventions. The included videos vary from 6 to 48 minutes, totaling approximately 13 hours of surgical footage. We labeled the data with surgical phase labels in close collaboration with clinical experts to ensure clinical validity and high data quality. Each surgery is segmented into five distinct phases: Preparation, Diagnosis, Femoral Tunnel Creation, Tibial Tunnel Creation, and ACL Reconstruction, with labels assigned at one label per second as detailed in Table \ref{tab:acl_phases}.

\begin{figure}[h!]
    \centering
    \begin{subfigure}[t]{0.18\textwidth}
        \includegraphics[width=\textwidth]{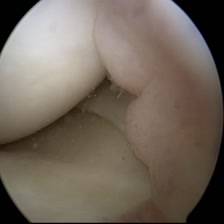}
    \end{subfigure}
    \hfill
    \begin{subfigure}[t]{0.18\textwidth}
        \includegraphics[width=\textwidth]{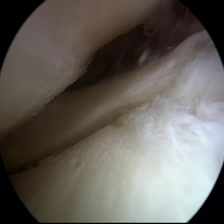}
    \end{subfigure}
    \hfill
    \begin{subfigure}[t]{0.18\textwidth}
        \includegraphics[width=\textwidth]{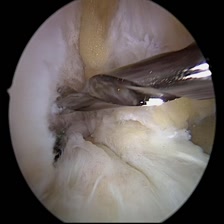}
    \end{subfigure}
    \hfill
    \begin{subfigure}[t]{0.18\textwidth}
        \includegraphics[width=\textwidth]{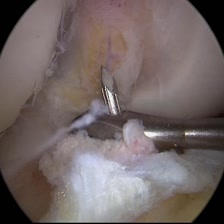}
    \end{subfigure}
    \hfill
    \begin{subfigure}[t]{0.18\textwidth}
        \includegraphics[width=\textwidth]{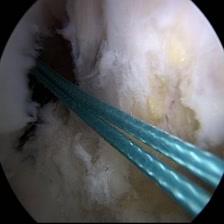}
    \end{subfigure}
    
    \vspace{1em}
    
    \begin{subfigure}[t]{0.18\textwidth}
        \includegraphics[width=\textwidth]{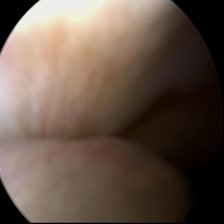}
        \caption*{Preparation}
    \end{subfigure}
    \hfill
    \begin{subfigure}[t]{0.18\textwidth}
        \includegraphics[width=\textwidth]{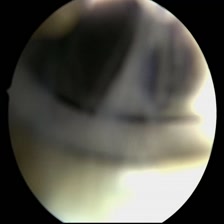}
        \caption*{Diagnosis}
    \end{subfigure}
    \hfill
    \begin{subfigure}[t]{0.18\textwidth}
        \includegraphics[width=\textwidth]{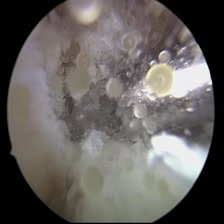}
        \caption*{Femoral Tunnel}
    \end{subfigure}
    \hfill
    \begin{subfigure}[t]{0.18\textwidth}
        \includegraphics[width=\textwidth]{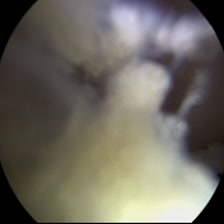}
        \caption*{Tibial Tunnel}
    \end{subfigure}
    \hfill
    \begin{subfigure}[t]{0.18\textwidth}
        \includegraphics[width=\textwidth]{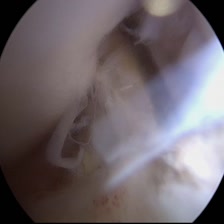}
        \caption*{ACL Reko}
    \end{subfigure}
    
    \caption{Top row: Clear images from each phase of ACL surgery, showcasing identifiable stages of the procedure. Bottom row: Corresponding unclear or blurry frames, demonstrating the challenge of phase recognition in arthroscopy. This contrast highlights the importance of advanced recognition techniques to improve accuracy in surgical training and real-time assistance.}
    \label{fig:phases}
\end{figure}

Fig. \ref{fig:phases} illustrates frames belonging to the respective phases. The figure also highlights the specific challenges of recognizing phases in arthroscopic ACL surgery videos compared to laparoscopic procedures like those in the Cholec80 dataset. Arthroscopy involves a smaller and more confined field of view, and the presence of fluids causes light reflection and refraction, creating visual distortions. Precise instrument navigation in narrow joint spaces increases the likelihood of occlusions, and the camera's proximity to the surgical site necessitates frequent adjustments. Blood and debris can quickly cloud the view, and variable lighting conditions lead to glare and shadows. The similar visual textures of joint structures make tissue differentiation difficult, and any movement can cause significant video disruptions. These challenges necessitate advanced recognition techniques to improve accuracy in surgical training and real-time assistance.

\begin{table}[h!]
    \centering
    \begin{tabular}{|l|c|c|c|}
        \hline
        \textbf{Phase} & \textbf{Total Duration} & \makecell{\textbf{\# of Videos} \\ \textbf{Containing Phase}} & \makecell{\textbf{Average Phase} \\ \textbf{Length}} \\
        \hline
        Preparation & 50\textquotesingle :40\textquotedbl & 19 & 2\textquotesingle :40\textquotedbl \\
        Diagnosis & 208\textquotesingle :32\textquotedbl & 25 & 8\textquotesingle :20\textquotedbl \\
        Femoral Tunnel Creation & 265\textquotesingle :54\textquotedbl & 26 & 10\textquotesingle :14\textquotedbl \\
        Tibial Tunnel Creation & 214\textquotesingle :02\textquotedbl & 25 & 8\textquotesingle :34\textquotedbl \\
        ACL Reconstruction & 54\textquotesingle :38\textquotedbl & 16 & 3\textquotesingle :25\textquotedbl \\
        \hline
    \end{tabular}

    \vspace{5pt} 

    \caption{This table provides a detailed breakdown of the ACL surgery dataset, highlighting the duration and distribution of the different surgical phases.}
    \label{tab:acl_phases}
\end{table}

\subsection{Temporal-aware Frame-wise Feature Extraction}
The detection of the surgical phases requires the analysis of a longer temporal context of surgical video, as high-level surgical phases are composed of several surgical actions, tool usage, and sub-tasks. However, the processing of long sequences of video data introduces a high computational complexity. Therefore, in previous work, dimensionality reduction techniques, such as the extraction of intermediate features as a preprocessing step, have been introduced \cite{2.twinanda2016endonet,czempiel2021opera}. While initially, frame-based feature extraction techniques were employed \cite{10_jin2017sv}, the integration of temporal context in the feature extraction stage has been shown to lead to higher-quality features and resulting performance in the downstream task of surgical phase prediction \cite{liu2023lovit}.

\begin{figure}[h!]
    \captionsetup{belowskip=0pt, aboveskip=5pt}
    \centering
    \includegraphics[width=\textwidth, trim=5 0pt 5pt 0pt, clip]{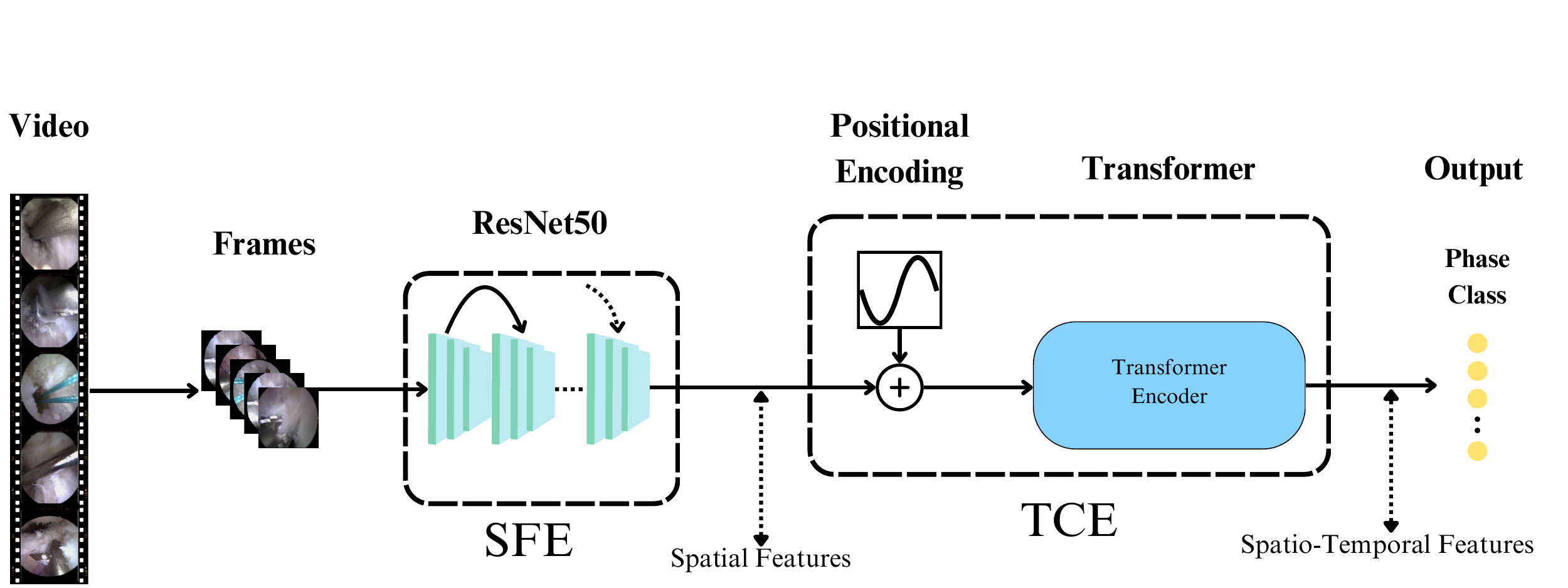}
    \caption{Architecture of our spatio-temporal feature extractor showcasing the process from initial feature extraction using ResNet-50 to enhanced temporal resolution via a transformer layer. The Spatial Feature Encoder (SFE) and Temporal Context Encoder (TCE) further analyze and integrate these features to produce comprehensive spatio-temporal vectors, critical for accurate surgical phase prediction.}
    \label{fig:architecture}
\end{figure}

Our model, as illustrated in Fig. \ref{fig:architecture}, takes a novel approach by emphasizing temporal-aware feature extraction from the outset. Unlike traditional methods that rely solely on frame-based feature extraction, our approach begins with a ResNet-50 model to perform initial feature extraction on three-channel images of size 240x240 pixels. This step generates a 2048-dimensional feature vector for each frame. However, we further enhance these features by employing a transformer layer with 2 heads to process and refine the extracted features, thus ensuring a deeper integration of temporal data. This dual approach of combining spatial features with temporal context right from the feature extraction phase enables our model to capture and utilize the sequential dependencies within the video data more effectively, enhancing the precision and accuracy of surgical phase recognition.

Building upon this enhanced feature set, we utilize a Spatial Feature Encoder (SFE) and a Temporal Context Encoder (TCE), as illustrated in figure \ref{fig:architecture}, to extract features that take the sequential nature of video data into account. The SFE transforms each video frame, denoted as \(x_t \in \mathbb{R}^{H \times W \times C}\), into a spatial feature vector. These spatial features are then compiled into a sequence for a given window of frames \(n\), creating a contiguous representation of the surgical procedure over time. The TCE processes this sequence to merge the spatial information with the temporal context, outputting a comprehensive spatio-temporal feature vector. 

\subsection{Surgical Progress Index}
\begin{figure}[h!]
    \centering
    \includegraphics[width=\textwidth, trim=150pt 200pt 150pt 50pt, clip]{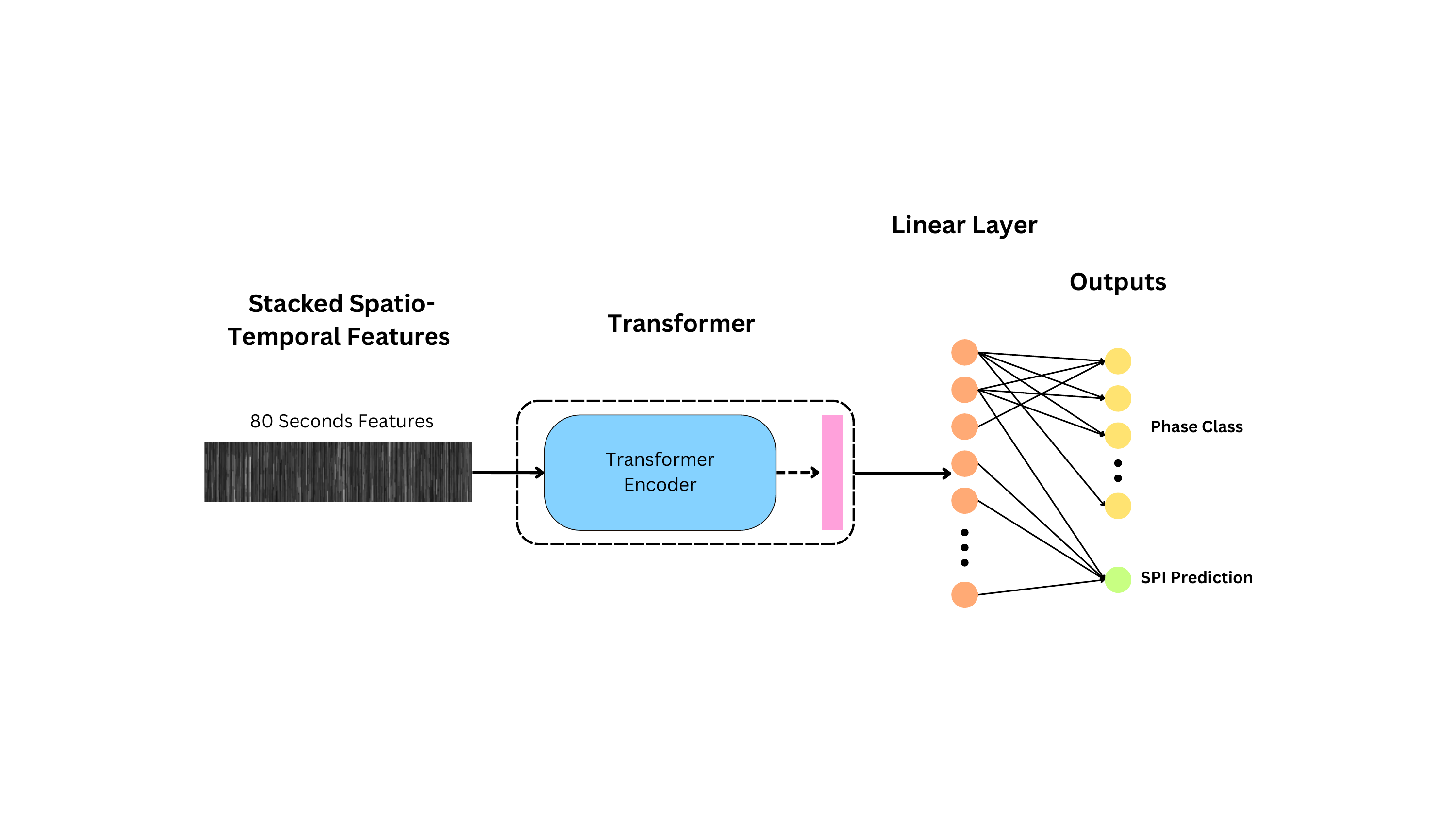}
    \caption{Overview of the final model for surgical phase recognition. The architecture has a transformer branch which processes an 80-second sequence of features. These features are derived from our spatio-temporal feature extractor, which combines spatial information from ResNet-50 with temporal context through transformer layers.}
    \label{fig:architecture2}
\end{figure}

Traditionally, surgical phase recognition models classify discrete phases without a continuous understanding of time. However, the progression of surgery is inherently sequential and can benefit from a regression-based approach, where understanding the flow of each phase is crucial. To enhance the capabilities of our phase recognition model, we introduce the Surgical Progress Index (SPI), a novel metric designed to quantify the progression of the surgery. The SPI addresses this by providing a continuous measure that enables the model to implicitly incorporate the global temporal context of the surgical procedure. This ability to integrate global context without processing the entire video sequence makes the SPI particularly valuable. It allows the model to make more accurate predictions and offers a deeper insight into the surgical phase by bridging the gap between mere classification and holistic temporal analysis, thus enhancing the precision and utility of our surgical phase recognition system. The overall architecture of the proposed model, including the learned features and the SPI component, is illustrated in Fig. \ref{fig:architecture2}.

rgical Progress Index (SPI) can be calculated in a straightforward manner, requiring minimal additional labelling effort. For a given surgery recording $k$, $t$ represents the elapsed time since the beginning of the surgery, which corresponds to the frame index at the last frame of the current sequence, considering that the video is recorded at a rate of 1 frame per second. \( T_k \) represents the entire duration of the surgery, corresponding to the total number of frames in the recording. Finally, the SPI is calculated as:
\[
SPI(t,k) = \frac{t}{T_k}
\]
This ratio provides a continuous measure, indicating the percentage completion of the surgery at any given time step. However, variations in surgical practices and recording methods can lead to missing phases in videos, potentially misleading the model’s predictions; to address this, we use only videos that contain all predefined surgical phases for baseline computation (12 videos in ACL27 and 65 in Cholec80). For surgeries with missing phases, we establish an average progression percentage for each phase using the complete videos. Table \ref{tab:spi_transitions} gives an overview of the statistical distribution of the coverage of phases throughout the dataset. These statistics are used to compute an average duration for missing phases.  The calculation for the SPI of a frame \( t \) in a video \( n \), which might have missing initial or final phases, is given by:

\[
SPI(t,k,i)^{\text{adjusted}} = \frac{t}{T_k} + \sum_{i \in \text{missing phases}} \text{avg\_SPI}(i) \cdot I(\text{phase } i \text{ is missing})
\]

where the indicator function \(I(\text{phase } i \text{ is missing})\) is defined as:

\[
I(\text{phase } i \text{ is missing}) = 
\begin{cases} 
1 & \text{if phase } i \text{ is missing in video } k \\
0 & \text{otherwise}
\end{cases}
\]

In this formula, \( t \) represents the frame index of the current sequence, and \( T_k \) denotes the total number of frames in the video. The term \(\text{avg\_SPI}(i)\) refers to the average SPI of the missing phase \( i \), which is determined from the analysis of videos that include all phases, as detailed in Table \ref{tab:spi_transitions}.

For instance, if a video starts from the second phase, the SPI for the first frame of this video is set to the average SPI of the first phase. This average SPI is calculated based on the analysis of videos containing all phases. By making this adjustment, the SPI more accurately reflects the surgical progress, thereby enhancing the model’s ability to generalize across various surgical recordings.

\begin{table}[h!]
    \centering
    \resizebox{\textwidth}{!}{
    \begin{tabular}{|c|c|c|c|c|c|c|c|c|}
        \hline
        & \makecell{\textbf{First} \\ \textbf{Frame}} 
        & \makecell{\textbf{First} \\ \textbf{Transition}} 
        & \makecell{\textbf{Second} \\ \textbf{Transition}} 
        & \makecell{\textbf{Third} \\ \textbf{Transition}} 
        & \makecell{\textbf{Fourth} \\ \textbf{Transition}} 
        & \makecell{\textbf{Fifth Transition} \\ \textbf{Last (ACL)}} 
        & \makecell{\textbf{Sixth} \\ \textbf{Transition}} 
        & \makecell{\textbf{Last Frame} \\ \textbf{(Cholec80)}} \\
        \hline
        \textbf{Cholec80} & 0.0 & 0.051 & 0.452 & 0.530 & 0.847 & 0.885 & 0.964 & 1.000 \\
        \hline
        \textbf{ACL27} & 0.0 & 0.073 & 0.309 & 0.534 & 0.765 & 1.000 & & \\
        \hline
    \end{tabular}
    }
    \vspace{5pt} 
    \caption{Average SPI transitions for Cholec80 and ACL27 datasets.}
    \label{tab:spi_transitions}
\end{table}

\subsection{Loss Function and Training Details}
For model training, we employ a composite loss function that integrates both phase classification and progress regression. Specifically, for the classification of surgical phases, we use Sparse Categorical Cross Entropy, and for SPI regression, we use Mean Absolute Error (MAE). The combined loss function is expressed as:
\[
L(p, y) = \lambda \left( -\sum_{i} y_i \log(p_i) \right) + (1 - \lambda) \left( \frac{1}{N} \sum_{i=1}^{N} |p_i - y_i| \right)
\]
where \( y_i \) is the ground truth label, \( p_i \) is the model's predicted probability for class \( i \), \( N \) is the number of samples, and \( \lambda \) is the weight assigned to each loss component. For our model, we set \( \lambda = 0.5 \), giving equal importance to both the classification accuracy and the regression precision.

All experiments were conducted using the TensorFlow 2.10.1 framework on an NVIDIA RTX-A6000 GPU. The model was trained using the Adam optimizer with an initial learning rate of 0.000005, scheduled to decrease by a factor of ten after every 10 epochs, over a total of 30 epochs. The batch size for training was set at 32. For the cross-validation procedure, the datasets were randomly divided into five equal partitions for both the ACL27 and Cholec80 datasets. In this context, the Cholec80 dataset was split such that 40 videos were allocated for training and testing the model's performance, respectively. Similarly, the ACL27 dataset was divided with 18 videos used for training and the remaining 9 videos reserved for validation. For the experiments not involving cross-validation, the data split was executed in the following manner. This approach was adopted to facilitate comparison with the majority of other studies in this domain, which typically followed the same methodology. Consequently, for the Cholec80 dataset, the first 40 videos were selected for training, and for the ACL27 dataset, the first 18 videos were chosen for training purposes. Results marked with an uppercase plus (+) indicate that they were obtained using the non-cross-validation data split.

\section{Results}
\subsection{Evaluation Metrics}
To determine the effectiveness of our surgical phase recognition model, we use a combination of standard metrics: Accuracy, Precision, Recall, and Jaccard Index. Accuracy evaluates the model's ability to correctly identify surgical phases in videos, irrespective of their duration. Given the imbalanced nature of our dataset, with some phases being significantly shorter than others, we also focus on average Precision, Recall, and Jaccard Index for a more detailed evaluation. These metrics collectively quantify how well our model predicts surgical phases and provide a comprehensive view of the model’s performance at both the video and phase level without being influenced by phase length variations. In previous research, some studies used cross-validation \cite{22_czempiel2020tecno,czempiel2021opera} while others did not and instead reported sample-based standard deviation (STD) \cite{2.twinanda2016endonet,18_jin2020multi, liu2023skit}. To enable comparison with all these studies, we evaluated both.

\subsection{Comparative Analysis with Existing Models}
To evaluate the performance of our model against existing methods, we implemented and trained Trans\_SVNet, MTRCNet, TeCNO, and Opera on the ACL27 dataset. The results based on the evaluation without cross-validation are summarized in Table \ref{tab:comparison_results}. Our model achieved an accuracy 72.91\%, a precision of 72.86\%, and a Jaccard Index of 57.39\%. In comparison, Trans\_SVNet models, evaluated with sequence lengths of 30, 60, and 80, achieved accuracies of 66.62\%, 66.75\%, and 66.73\%, respectively, with corresponding Jaccard Indices of 47.96\%, 48.05\%, and 48.03\%. MTRCNet and TeCNO reported lower performance, with MTRCNet achieving an accuracy of 58.75\% and TeCNO achieving 58.31\%. This comparative analysis shows that our model outperforms existing methods on the new ACL27 dataset, particularly in terms of accuracy and precision.

\begin{table}[h!]
    \centering
    \begin{tabular}{|l|c|c|c|c|}
        \hline
        \textbf{Model} & \textbf{Accuracy (\%)} & \textbf{Precision (\%)} & \makecell{\textbf{Jaccard} \\ \textbf{Index (\%)}} & \textbf{F1-Score (\%)} \\
        \hline
        Trans\_SVNet\textsuperscript{+}~\cite{gao2021trans} (Seq = 30s) & 66.62 & 66.51 & 47.96 & - \\
        Trans\_SVNet\textsuperscript{+}~\cite{gao2021trans} (Seq = 60s) & 66.75 & 66.45 & 48.05 & - \\
        Trans\_SVNet\textsuperscript{+}~\cite{gao2021trans} (Seq = 80s) & 66.73 & 66.43 & 48.03 & - \\
        MTRCNet\textsuperscript{+}~\cite{18_jin2020multi} (Seq = 4s) & 58.75 & 59 & - & 58 \\
        MTRCNet\textsuperscript{+}~\cite{18_jin2020multi} (Seq = 10s) & 60.68 & 63 & - & 59 \\
        TeCNO\textsuperscript{+}~\cite{22_czempiel2020tecno} & 58.31 & 66.05 & 40.66 & - \\
        \textbf{Ours\textsuperscript{+}} & \textbf{72.91} & \textbf{72.86} & \textbf{57.39} & - \\
        \hline
    \end{tabular}
    \vspace{5pt} 
    \caption{Comparison of performance metrics for different models on the ACL27 dataset.}
    \label{tab:comparison_results}
\end{table}

These results highlight the model's capability to accurately classify surgical phases and predict the progression of surgery. In addition to phase recognition, the model's performance in predicting the Surgical Progress Index (SPI) was evaluated. The SPI, which quantifies the progression of the surgery on a scale from 0 to 1, demonstrated an output error of 10.6\%. This low error rate indicates the model's effectiveness in providing a reliable estimate of the surgery's progression, thus enabling a more continuous and nuanced understanding of the surgical workflow.

\subsection{Ablation Study}

To thoroughly understand the contributions of the proposed components in our model, we conducted an ablation study. This study evaluates the inclusion of the Surgical Progress Index (SPI) and the spatio-temporal features on the performance of our model. The results of different configurations are summarized in Table \ref{tab:ablation_study}. The results were obtained using a 5-round cross-validation approach to ensure robustness and reliability of the findings.

\begin{table}[h!]
    \centering
    \begin{tabular}{|c|c|c|c|c|c|}
        \hline
        \makecell{\textbf{Spatio-temporal} \\ \textbf{Features}} & 
        \textbf{SPI} &
        \textbf{Accuracy} &
        \textbf{Precision} &
        \textbf{Recall} &
        \makecell{\textbf{Jaccard} \\ \textbf{Index}} \\
        \hline
        \ding{55} & \ding{55} & $66.38 \pm 1.71$ & $66.818 \pm 2.26$ & $65.48 \pm 2.59$ & $48.92 \pm 2.90$ \\
        \ding{55} & \ding{51} & $69.83 \pm 3.39$ & $69.67 \pm 3.66$ & $69.32 \pm 3.20$ & $53.19 \pm 3.82$ \\
        \ding{51} & \ding{55} & $71.23 \pm 2.96$ & $72.58 \pm 2.26$ & $71.12 \pm 3.41$ & $55.79 \pm 4.02$ \\
        \ding{51} & \ding{51} & $\mathbf{76.71 \pm 2.44}$ & $\mathbf{76.44 \pm 3.39}$ & $\mathbf{76.13 \pm 3.76}$ & $\mathbf{62.184 \pm 4.44}$ \\
        \hline
    \end{tabular}
    \vspace{5pt} 
    \caption{Performance Metrics for Different Model Configurations on ACL Dataset}
    \label{tab:ablation_study}
\end{table}

This table summarizes the impact of incorporating the spatio-temporal features and the SPI on the model's performance across various metrics, namely Accuracy, Precision, Recall, and the Jaccard Index. A cross (\ding{55}) indicates the absence of a component, while a check mark (\ding{51}) denotes its presence. The first row of the table shows the baseline model, which lacks both spatio-temporal features and SPI. This configuration yields the lowest performance, with an accuracy of $66.38 \pm 1.71$, precision of $66.818 \pm 2.26$, recall of $65.48 \pm 2.59$, and a Jaccard Index of $48.92 \pm 2.90$. When spatio-temporal features are added (second row), there is a notable improvement across all metrics. The accuracy increases to $71.23 \pm 2.96$, precision to $72.58 \pm 2.26$, recall to $71.12 \pm 3.41$, and the Jaccard Index to $55.79 \pm 4.02$. The inclusion of only the SPI (third row) also enhances the model's performance compared to the baseline. The accuracy is $69.83 \pm 3.39$, precision is $69.67 \pm 3.66$, recall is $69.32 \pm 3.20$, and the Jaccard Index is $53.19 \pm 3.82$. The final configuration, which incorporates both spatio-temporal features and the SPI (fourth row), achieves the best performance across all metrics. The accuracy is the highest at $76.71 \pm 2.44$, precision at $76.44 \pm 3.39$, recall at $76.13 \pm 3.76$, and the Jaccard Index at $62.184 \pm 4.44$.

\subsection{Benchmark on Cholec80 Dataset}
We also benchmarked our proposed method on the standard Cholec80 dataset to compare its performance with previous work. This dataset has been extensively utilized in previous research for surgical phase recognition, making it an ideal benchmark for evaluating our model. Table \ref{tab:benchmark_cholec80} presents a comparative analysis of our model with the other state-of-the-art methods. The table is divided into two parts: the top section lists results from methods using the first 40 videos for training and the remaining 40 for validation, while the bottom section lists results from methods using cross-validation. For each approach, we also report our model's performance using the corresponding data splits.

\begin{table}[h!]
    \centering
    \begin{tabular}{|l|r|r|r|}
        \hline
        \textbf{Method} & \textbf{Accuracy} & \textbf{Precision} & \textbf{Recall}  \\
        \hline   
        EndoNet\textsuperscript{+}~\cite{2.twinanda2016endonet} & $81.7 \pm 4.2$ & $73.7$ & $79.6$\\
        PhaseNet\textsuperscript{+}~\cite{2.twinanda2016endonet} & $78.8 \pm 4.7$ & $71.3$ & $76.6$ \\
        MTRCNet\textsuperscript{+}~\cite{18_jin2020multi} & $89.2 \pm 7.6$ & $86.9$ & $88.0$ \\
        LoViT\textsuperscript{+}~\cite{liu2023lovit} & $91.5 \pm 6.1$ & $83.1$ & $86.5$ \\
        SKiT\textsuperscript{+}~\cite{liu2023skit} & \textbf{92.5 $\pm$ 5.1} & $84.6$ & \textbf{88.5} \\
        \textbf{Ours}\textsuperscript{+} & \textbf{92.4 $\pm$ 5.0} & \textbf{85.6} & $80.6$ \\
        \hline
        TeCNO~\cite{22_czempiel2020tecno} & $88.5 \pm 0.27$ & $81.6$ & $85.2$ \\
        OperA~\cite{czempiel2021opera} & $91.2 \pm 0.6$ & $82.1$ & \textbf{86.92} \\
        \textbf{Ours} & \textbf{91.9 $\pm$ 0.9} & \textbf{83.1} & $86.5$ \\
        \hline
    \end{tabular}
    \vspace{5pt} 
    \caption{Benchmark Performance Comparison on Cholec80 Dataset}
    \label{tab:benchmark_cholec80}
\end{table}

Our model demonstrates competitive performance with an accuracy of \(92.4 \pm 5.0\%\), which is very close to SKiT's highest accuracy of \(92.5 \pm 5.1\%\). Our model achieves a precision of 85.6, higher than both SKiT (84.6) and LoViT (83.1), and a recall of 80.6. In the bottom part, we compare methods that employed cross-validation. Our model achieves the highest accuracy (\(91.9 \pm 0.9\%\)), and precision (83.1) demonstrating its robustness and reliability across different evaluation schemes.

In addition to these metrics, we also evaluated the model's performance in predicting the Surgical Progress Index (SPI) on the Cholec80 dataset. The SPI, which quantifies the progression of the surgery on a scale from 0 to 1, demonstrated an output error of $9.86\%$.

\section{Discussion}
Automated workflow analysis and surgical phase recognition hold significant potential for improving orthopedic surgery by enhancing precision and workflow efficiency. While these technologies have been effectively applied in various surgical domains to aid real-time decision-making and documentation, their use in arthroscopy has been limited. Arthroscopic procedures, commonly employed in orthopedics, present unique challenges such as limited visual fields and frequent occlusions, which complicate phase recognition. By addressing these challenges, advanced phase recognition systems can improve surgical training, assist surgeons during procedures, and streamline clinical workflows. This study introduces the ACL27 dataset and a novel transformer-based model designed to meet the specific needs of arthroscopic video analysis, providing a valuable resource and a new benchmark for future research in this area.

Our approach addresses these challenges by enriching the feature extraction process with temporal information directly integrated into the spatial feature extraction stage. This methodology enhances the model's ability to discern subtle transitions and similar visual appearances across different phases, which are critical in arthroscopic environments.

The integration of the Surgical Progress Index (SPI) enhances the model's capability to not only recognize but also predict the progression of surgery, an aspect crucial for real-time surgical assistance systems. The SPI provides a continuous measure of surgery progression, incorporating global temporal context, which is essential for accurately mapping surgical dynamics and predicting the remaining duration of surgery. This capability is particularly beneficial for operating room (OR) planning and management. By providing an estimate of the surgery's remaining duration, the SPI can help optimize OR scheduling, reduce patient wait times, and improve overall surgical efficiency.

Empirical results from testing our model on the newly introduced ACL27 dataset and the established Cholec80 dataset demonstrate its robustness in handling complex surgical phase recognition tasks. Our proposed method and dataset for ACL phase recognition highlight the model's adaptability to different surgical environments and its resilience against variations inherent in live surgical scenarios. The model achieved an accuracy of $72.91 \pm 11.76\%$ on the ACL27 dataset. Since recent works on other types of surgeries like cholecystectomy cannot be compared on ACL due to unavailable code, we benchmarked our model on the Cholec80 dataset, achieving an accuracy of $92.4 \pm 5.0\%$, which is comparable to existing methods.

One of the key contributions of our work is the potential clinical impact. By enabling more accurate and timely recognition of surgical phases, our model can support real-time decision-making, potentially reducing surgery times and improving patient outcomes. The SPI, in particular, offers significant practical benefits for OR management. Accurate predictions of surgical progress and remaining duration can facilitate better resource allocation, improve patient flow, and reduce the likelihood of scheduling conflicts.

While our results are promising, several limitations warrant further discussion. Firstly, the size of the ACL27 dataset, although sufficient for initial validation, could be expanded to improve generalizability. Standardizing recording protocols is crucial, as variations in camera placement, lighting, and occlusions can affect phase recognition accuracy. Additionally, the issue of missing phases in some surgical videos can mislead predictions and impact the SPI calculation. Although we used average progression percentages for missing phases, more sophisticated methods are needed. Labeling efforts pose another significant limitation. Accurate annotation requires significant expertise and is labor-intensive, introducing potential human error and variability. Developing automated or semi-automated labeling tools could streamline this process and enhance label accuracy. The computational demands of our model are substantial, potentially limiting its practical applicability. Future work should focus on optimizing the model’s efficiency to make it more accessible for real-time surgical assistance.

\section{Discussion}
\paragraph{Overall Accuracy}

Our proposed model achieved a notable accuracy of $72.91 \pm 11.76\%$ on the ACL27 dataset, specifically designed for arthroscopic surgery, and $92.4 \pm 5.0\%$ on the widely-used Cholec80 dataset. These results underscore the effectiveness of our approach across different surgical domains. On the ACL27 dataset, our model significantly outperformed existing methods. For instance, when compared to other models adapted to this dataset, our model's accuracy of 72.91\% is considerably higher than the 66.75\% accuracy achieved by the Trans-SVNet model with an 80-second sequence length. Similarly, MTRCNet and TeCNO, two other prominent models in surgical phase recognition, reported lower accuracies of 60.68\% and 58.31\%, respectively. These comparisons, as detailed in Table 3 of our paper, highlight our model's superior capability to handle the specific challenges of arthroscopic video analysis, such as limited fields of view and frequent occlusions. Moreover, our model's performance on the Cholec80 dataset, with an accuracy of 92.4\%, is comparable to other leading methods in the field. For example, SKiT and LoViT achieved accuracies of 92.5\% and 91.5\%, respectively. This demonstrates that our approach not only excels in the specialized context of arthroscopy but also maintains strong performance in more general surgical phase recognition tasks.

\paragraph{Feature Extractor}

The feature extraction process plays a critical role in the overall performance of surgical phase recognition models. In earlier works, such as the OperA model \cite{czempiel2021opera}, only spatial features were extracted, which, while effective, limited the model's ability to capture temporal dependencies essential for surgical video analysis. The LoViT model \cite{liu2023lovit} made significant strides by incorporating a temporally-rich feature extractor with a temporal aggregator. However, during the final feature extraction step, LoViT reverted to extracting only spatial features, leaving temporal dependencies less integrated into the final feature representation. Our approach advances this by directly incorporating spatio-temporal features into the feature extraction process. The effectiveness of this approach is clearly demonstrated in our ablation study (Table 4). When comparing models trained without spatio-temporal features (first two rows) to those that include them (last two rows), there is a marked improvement in performance. Specifically, the introduction of spatio-temporal features led to significant gains in accuracy, precision, recall, and Jaccard Index, highlighting the critical importance of integrating temporal information during feature extraction. Moreover, the use of spatio-temporal features has proven beneficial across various domains, including action recognition in videos \cite{carreira2017quo}, human activity recognition \cite{feichtenhofer2019slowfast}, and anomaly detection \cite{gong2019memorizing}, reinforcing its necessity for achieving high accuracy in complex tasks like surgical phase recognition.

\paragraph{Technical Aspects of SPI}
The Surgical Progress Index (SPI) is a novel contribution of our work that significantly enhances model performance by providing a continuous measure of surgery progression. By leveraging global temporal context, the SPI helps suppress incorrectly classified noisy frames, thereby improving overall accuracy, as demonstrated in our ablation study. This is particularly crucial given the high variability in surgery times and phase durations, which the SPI effectively manages, maintaining temporal coherence in phase predictions. Unlike other approaches, such as MTRC-Net \cite{18_jin2020multi}, which included additional outputs like tool detection that require extensive labeling efforts, our SPI offers a nearly effortless yet powerful enhancement to the model, making it both efficient and practical for clinical application, particularly in the context of arthroscopy.

\paragraph{Clinical Impact}
Our work presents a significant clinical impact by introducing the first tailored approach for surgical phase recognition in arthroscopy, making the procedure more quantifiable and measurable in an automated manner. This model can be integrated into video documentation systems for automated labeling and detailed workflow analysis, enhancing surgical training and real-time decision-making. The Surgical Progress Index (SPI) further contributes by offering accurate estimates of surgery duration, optimizing OR scheduling, and improving surgical efficiency. Moreover, this approach supports the development of more context-aware intelligent computer-assisted and robotic systems, as explored in recent studies on surgical guidance in complex procedures \cite{kolbinger2023artificial}.

\paragraph{Limitations}
Despite the promising results, several limitations should be acknowledged. The relatively small size of the ACL27 dataset, while sufficient for initial validation, limits the generalizability of our findings. Expanding this dataset would likely improve the robustness of the model, particularly in diverse clinical settings. The specific challenges of ACL surgeries, such as limited visual fields, frequent occlusions, and the presence of fluid and debris, further complicate phase recognition and may impact the model's accuracy. Additionally, variations in recording protocols, including camera placement, lighting, and surgeon technique, can affect the consistency of phase recognition. The issue of missing phases in some surgical videos also presents a challenge, particularly for SPI calculation, where more sophisticated methods could be explored to enhance accuracy.

Furthermore, the computational demands of our model, while necessary for achieving high accuracy, may limit its practical applicability in real-time surgical environments. Future work should focus on optimizing the model’s efficiency and robustness to ensure its feasibility for integration into clinical workflows and its adaptability across different surgical contexts.

\section{Conclusion and Outlook}

This study introduces a novel approach to surgical phase recognition in arthroscopic procedures, focusing on ACL reconstruction. We present the ACL27 dataset, which fills a critical gap in publicly available data for arthroscopy research. Our transformer-based model, leveraging temporal-aware frame-wise feature extraction demonstrates significant improvements in phase recognition accuracy and temporal awareness. The introduction of the Surgical Progress Index (SPI) provides a continuous measure of surgical progress, enhancing the model's practical applicability in real-time surgical assistance and planning. Experimental results on the ACL27 and Cholec80 datasets validate the robustness and generalizability of our approach. In conclusion, this study sets a new benchmark in surgical phase recognition for arthroscopic procedures and provides a robust framework for developing advanced surgical assistance systems, ultimately contributing to improved surgical outcomes and patient care.

\section{Acknowledgments}
This research is funded by the Innosuisse Flagship project PROFICIENCY No. PFFS-21-19. Additionally, this work has received support from the OR-X, a Swiss national research infrastructure for translational surgery, with associated funding provided by the University of Zurich and the University Hospital Balgrist.

\bibliographystyle{unsrt}

\end{document}